\documentclass[10pt,twocolumn,letterpaper]{article}

\usepackage{iccv}
\usepackage{times}
\usepackage{epsfig}
\usepackage{graphicx}
\usepackage{amsmath}
\usepackage{amssymb}

\usepackage[normalem]{ulem}
\usepackage{xcolor}

\usepackage[pagebackref=true,breaklinks=true,letterpaper=true,colorlinks,bookmarks=false]{hyperref}

\usepackage[capitalize]{cleveref}
\crefname{section}{Sec.}{Secs.}
\Crefname{section}{Section}{Sections}
\Crefname{table}{Table}{Tables}
\crefname{table}{Tab.}{Tabs.}

\newcommand{\tabincell}[2]{\begin{tabular}{@{}#1@{}}#2\end{tabular}}  
\definecolor{blue}{RGB}{0, 0, 255}
\newcommand{\cxj}[1]{\textcolor{blue}{#1}}

\iccvfinalcopy 

\def\iccvPaperID{16} 

\ificcvfinal\fi

\begin{document}

\title{SCSC: Spatial Cross-scale Convolution Module to Strengthen both CNNs and Transformers}

\author{Xijun Wang$^{1*}$, Xiaojie Chu$^{2*}$, Chunrui Han$^2$, Xiangyu Zhang$^2$\\
$^1$The University of Maryland, College Park, Maryland, USA. $^2$MEGVII Technology, Beijing, China\\
{\tt\small xijun@umd.edu, \{chuxiaojie, hanchunrui, zhangxiangyu\}@megvii.com}
}

\maketitle
\ificcvfinal\thispagestyle{empty}\fi

\begin{abstract}
This paper presents a module, Spatial Cross-scale Convolution (SCSC), which is verified to be effective in improving both CNNs and Transformers. Nowadays, CNNs and Transformers have been successful in a variety of tasks. Especially for Transformers, increasing works achieve state-of-the-art performance in the computer vision community. Therefore, researchers start to explore the mechanism of those architectures. Large receptive fields, sparse connections, weight sharing, and dynamic weight have been considered keys to designing effective base models~\cite{liu2021swin,han2021connection,zhao2021battle,rao2021global}. However, there are still some issues to be addressed: large dense kernels and self-attention are inefficient, and large receptive fields make it hard to capture local features. Inspired by the above analyses and to solve the mentioned problems, in this paper, we design a general module taking in these design keys to enhance both CNNs and Transformers. SCSC introduces an efficient spatial cross-scale encoder and spatial embed module to capture assorted features in one layer. On the face recognition task, FaceResNet with SCSC can improve \textbf{2.7\%} with \textbf{68\%} fewer FLOPs\footnote{The number of multiply-adds operations. $^*$Equal contribution. This work was completed when Xijun Wang was an intern at MEGVII.} and \textbf{79\%} fewer parameters. On the ImageNet classification task, Swin Transformer with SCSC can achieve even better performance with \textbf{22\%} fewer FLOPs, and ResNet with CSCS can improve \textbf{5.3\%} with similar complexity.  Furthermore, a traditional network (e.g., ResNet) embedded with SCSC can match Swin Transformer's performance. 



\end{abstract}
\vspace{-5mm}
\section{Introduction}
\label{sec:intro}

\begin{figure*}[h]
\begin{center}
\includegraphics[width=1.0\linewidth]{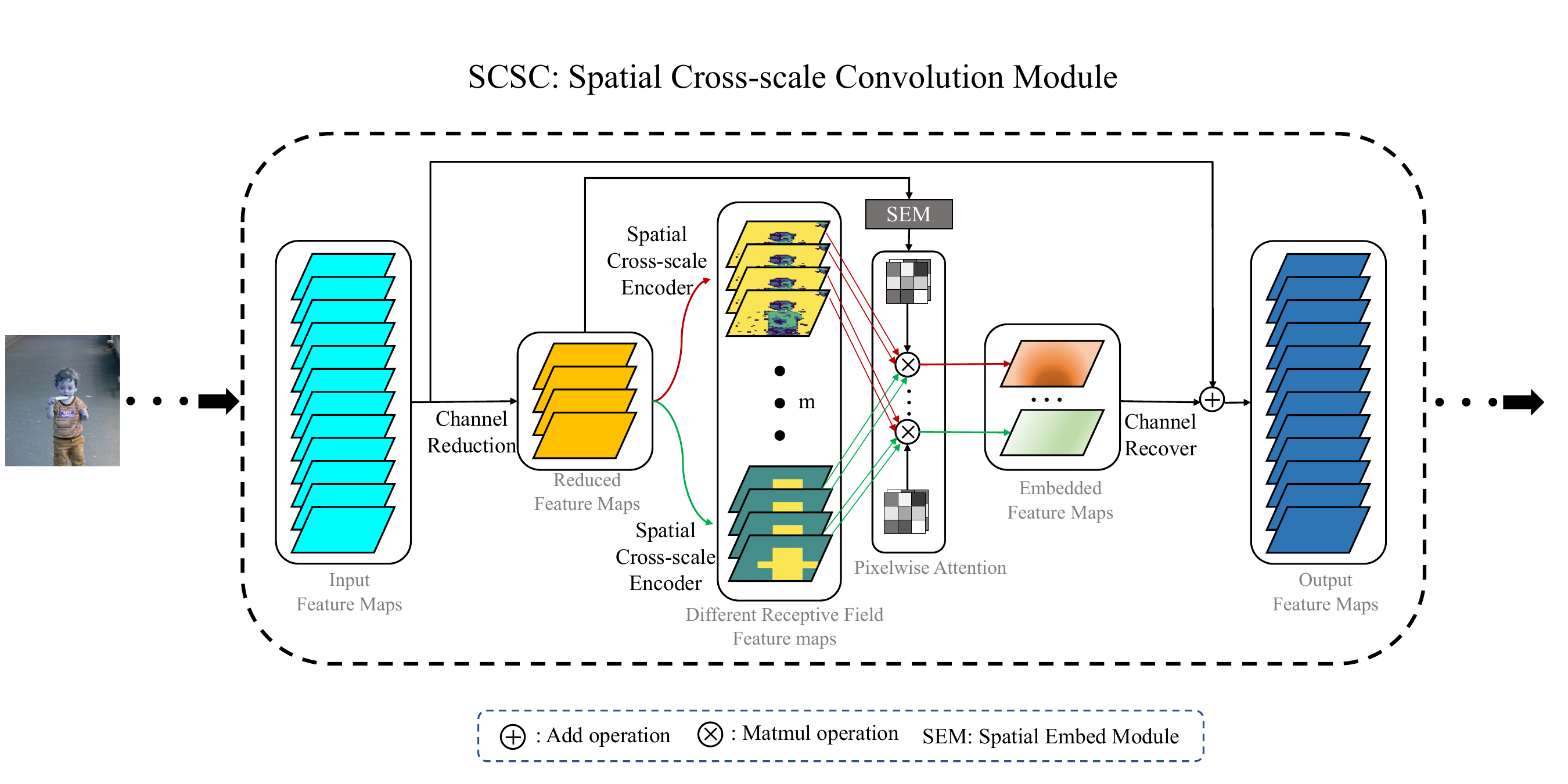}
\vspace*{-10mm}
\end{center}
\caption{Illustration of Spatial Cross-scale Convolution Module (SCSC) with different receptive fields in one layer. First, we decrease the input feature maps' channel to get the channel reduced features maps, which can save the channel-wise convolution computational cost. Second, we apply $m$ different Spatial encoder to get $m$ different receptive fields feature maps. Third, we use Spatial Embed Module to merge the $m$ feature maps to get the embedded feature maps. Finally, we recover the channel number as same as the input feature maps to get the output feature maps.}
\label{Fig:Introduction}
\vspace*{-5mm}
\end{figure*}

Vision transformers have achieved impressive breakthroughs in a variety of tasks, including classification, object detection, segmentation, video action recognition, and \etc\cite{dosovitskiy2020image,touvron2021training,liu2021swin}, making transformers promising backbones for vision applications. Transformers have a strong representation ability for their special designs, supporting a variety of data forms (tensor, set, sequence, graph, and \etc), and being robust to blocks and noise~\cite{cordonnier2019relationship,dong2021attention,xie2019feature,paul2021vision}. As a result, more and more researchers are trying to figure out what makes transformers so powerful. Han et al.~\cite{han2021connection} and ~\cite{zhao2021battle} analyze vision transformers from the respective of receptive fields, sparse connectivity, weight sharing, and dynamic weight, which has been considered as desired properties in model architecture design. 

For receptive field, \cite{dosovitskiy2020image,liu2021swin,dong2021cswin,szegedy2017inception} have shown that large kernels can efficiently introduce large receptive fields and can partially avoid the optimization problem caused by the increase of model depth. \cite{szegedy2017inception,dai2017deformable,chen2019detnas,peng2017large,chen2017deeplab} has proved that large kernel (kernel size $\ge 5\times5$) applied to CNN can obtain competitive performance, especially for downstream tasks. However, the intensive computational cost of large dense kernels makes it hard to be widely used in practice. For Transformer, multi-head self-attention layer can simulate a convolutional layer by linear projection operations ~\cite{cordonnier2019relationship}, to some extent, transformer acts like a global receptive field CNN network. But the high computational cost of self-attention still hinder transformers to be applied in practice.

To alleviate the intensive computational complexity,  \cite{liu2021swin} and \cite{han2021connection} have explored the sparse connectivity property, \cite{liu2021swin} proposed shifted windowing scheme by limiting self-attention computation to local windows. And \cite{han2021connection} utilizes $7\times7$ depth-wise convolution\cite{chollet2017xception,howard2017mobilenets} to simulate the local window scheme. However, the efficiency and spatial modeling ability still can be improved. Moreover, the receptive field in one layer could be more diverse.

For weight sharing, both CNN and Swin-Transformer \cite{liu2021swin} have applied weight sharing mechanism to obtain computational efficiency. The difference is CNNs share weights across spatial dimension while Swin-Transformer shares weight across channel dimension. 

For dynamic weight, there are many works ~\cite{hu2018squeeze,jia2016dynamic,yang2019condconv,chen2021dynamic} illustrate its effectiveness on CNN.  And transformers' self-attention structure uses dynamic weight for each input instance so that model capacity is increased. 

To conclude, all the four properties listed above in various architectures have potential to improve performance or efficiency. Convolution is efficient because of its spatial sharing pattern, and transformers have a large capacity because of their self-attention's large receptive field and dynamic scheme. Therefore, convolution and self-attention have complementary qualities, and a well-designed module that combines all desirable properties is possible.


To deal with all the above limitations and exploit complementary qualities of Transformer and CNN once for all, in this work, we propose a Spatial Cross-scale Convolution Module (SCSC), which can capture microscopic and macroscopic feature representation synchronously without intensive computational cost. As shown in Figure~\ref{Fig:Introduction}, for SCSC, different from the mainstream CNN's small kernel size or the Transformer's global receptive field, we design an intermediate expression for the receptive field and spatial modeling ability by using a wide range of kernel sizes from $3\times3$ to $13\times13$, depth-wise convolution is applied for its efficiency. Small kernels (e.g., $3\times3$) have advantage in modeling low-level and detailed information as shown in the upper red path in Figure~\ref{Fig:Introduction}, large kernels (e.g., $13\times13$) can handle the semantic dependence in a large receptive field as shown in the lower green path in Figure~\ref{Fig:Introduction}. Therefore, keeping both of them can obtain diverse spatial representations. Furthermore, we design an efficient spatial embed module to aggregate the different spatial representation features, which can capably integrate different levels of information. As a result, in our proposed SCSC, we can acquire different receptive fields in one layer, share weight in both spatial-wise and channel-wise, exploit depth-wise convolution to hold the sparse connectivity property, and apply the spatial embed module to bring dynamic connection across the channel dimension. 

Moreover, we have evaluated the proposed SCSC in different tasks. On ImageNet classification, Swin-T embedded with SCSC obtains 81.6\%  (VS Swin-T: 81.3\%) Top-1 accuracy with 22\% (1G) fewer FLOPs. ConvNet embedded with SCSC obtains 82.3\% Top-1 accuracy. On MS1M face recognition, FaceResNet embedded with SCSC achieves 95.6\% (VS FaceResNet: 92.6\%) rank-1 face identification accuracy with 68\% fewer FLOPs and 79\% fewer parameters . On COCO detection, Swin-SCSC with Mask R-CNN gains 43.2\% (VS Swin-T: 42.7) box AP with 23G fewer FLOPs. On ADE20K segmentation, ResNet-SCSC achieve 45.7\% mIoU (VS Swin-T: 44.4\%). These experiments demonstrate the effectiveness of our proposed SCSC.





\noindent To summarize, we make the following contributions: \\
\textbf{1) }Present a high capacity and effective convolution module SCSC, which can dynamically combine a large range of receptive fields (\textbf{pixel-wise}) in one layer to enhance the presentation ability. \\
\textbf{2) } Architectures applied with SCSC can obtain better performance with fewer computational cost and parameters. Furthermore, SCSC is a general module and can be applied to strengthen both CNNs and Transformers. \\
\textbf{3) }SCSC module can power the classical neural networks (e.g., ResNet50) to achieve comparable performance with strong Transformers (e.g., Swin).
    

\section{Related Work}
\label{sec:related_work}

\subsection{Vision Transformers}
Recently, increasing Vision Transformers obtain state-of-the-art performance on visual tasks \cite{khan2021transformers,han2020survey}.
To improve the original vision Transformer (ViT)~\cite{dosovitskiy2020image}, \cite{chu2021conditional} offers a conditional positional encoding (CPE) technique. Unlike prior fixed or learnable positional encodings, which are pre-defined and independent of input tokens, CPE is dynamically produced and conditioned by the immediate neighborhood of the input tokens. For self-attention, \cite{d2021convit} provides gated positional self-attention (GPSA), a kind of positional self-attention that includes a "soft" convolutional inductive bias. And \cite{guo2021beyond} presents a new attention mechanism called external attention, which substitutes self-attention in existing popular architectures. External attention is linear in complexity and implicitly considers all data samples' correlations.  By preserving encoder branches at various scales while engaging attention across scales, \cite{wu2021cvt} introduces a co-scale mechanism to image Transformers. \cite{yuan2021tokens} creates a progressive tokenization system in order to overcome the restriction of ViT when training from scratch on a medium-sized dataset such as ImageNet.

To make Transformer more efficient, \cite{wu2021cvt} introduces Convolutional Vision Transformer (CvT), which enhances the performance and efficiency of Vision Transformer (ViT) by incorporating convolutions into ViT to provide the best of both designs. \cite{liu2021swin} proposes a hierarchical Transformer named Swin Transformer that computes its representation using Shifted windows. The shifted windowing scheme brings greater efficiency by limiting self-attention computation to non-overlapping local windows while also allowing for cross-window connection, making Swin an efficient and effective Vision Transformer architecture. To go further, \cite{dong2021cswin} creates CSWin Self-Attention, which divides multi-heads into parallel groups and conducts self-attention in horizontal and vertical stripes. 

For different transformers design and learning strategies, \cite{touvron2021training} constructs competitive convolution free transformers DeiTs, which can compete with the state-of-the-art on ImageNet without using any external data at that time. They also present a transformer-specific teacher-student method. \cite{wang2021pyramid} offers Pyramid Vision Transformer (PVT), which addresses the challenges of applying Transformer to a variety of dense prediction applications. Moreover, CrossFormer~\cite{wang2021crossformer} uses cross-scale convolution as a downsample (embedding) layer which blends each embedding with multiple patches of different scales for self-attention module.

Instead of proposing algorithms only to complement the shortcomings of ViTs, our method absorbs the advantages of vision transformer (e.g. large receptive field) and go a further step to benefit both transformer and CNN.

\subsection{Large Kernel and Spatial Modeling}

In the exploration of large kernels, DetNAS \cite{chen2019detnas} chooses large-kernel blocks in low-level layers and deep blocks in high-level layers. \cite{peng2017large} discovers that the large kernel (and effective receptive field) plays a crucial role when we have to do classification and localization tasks at the same time (e.g., semantic segmentation). To solve both classification and localization challenges in semantic segmentation, \cite{peng2017large} presents a Global Convolutional Network. Deeplab~\cite{chen2017deeplab} applies "atrous convolution" with upsampled filters for dense feature extraction for semantic segmentation and expands it even further to atrous spatial pyramid pooling, which stores objects and visual information at several scales. ConvNet~\cite{liu2022convnet} revisits the use of large kernel-sized ($7 \times 7$) convolutions and RepLKNet~\cite{ding2022scaling} further scale up receptive fields using $31 \times 31$re-parameterized large depth-wise convolutions. However, RepLKNet~\cite{ding2022scaling} focus on large models with a large number of parameters ($\ge$ 79M). 

To better model spatial information, \cite{dai2017deformable} presents deformable ConvNets for modeling dense spatial transformation to learn receptive fields adaptively. 
Inception family (e.g., GoogLeNet~\cite{szegedy2015going} and Inception-V4 \cite{szegedy2017inception}) extract multi-scale features by different convolutional kernels and fuse them statically using concatenation. DRConv \cite{chen2021dynamic} learns a guided mask to assign different customized weights to different spatial regions for better spatial representation. SKNet~\cite{li2019selective} proposes a dynamic selection technique in CNNs that allows each neuron to modify its receptive field size adaptively based on different scales of input.


Compared with these methods, we additionally introduce large range of receptive field (especially large receptive field), and consider weight sharing and dynamic mechanism at the same time.

\subsection{Dynamic Mechanism}

With the prevalence of data dependency mechanism~\cite{allport1989visual,jaderberg2015spatial,vaswani2017attention} , which emphasizes to extract more customized feature for diverse representation~\cite{mahajan2018exploring}.  Benefited from the data dependency mechanism, networks can flexibly adjust themselves, including the structure and parameters, to automatically fit the fickle information to improve the representation ability of neural networks. \cite{chen2017sca,woo2018cbam} indicate that different regions in the spatial dimension are not equally important in representation learning and should be processed differently. For instance, activation in important regions needs to be amplified to play a dominant role in the forward propagation. SKNet~\cite{li2019selective} designs a dynamic module to channel-wisely select suitable receptive fields based on channel attention and achieves better performance. It dynamically restructures the networks for the sake of different receptive fields with dilated convolutions~\cite{yu2015multi,yu2017dilated}.




From the aspect of dynamic weights, CondConv~\cite{yang2019condconv} obtains dynamic weights by the dynamical linear combination of several weights. 
And the specialized convolution kernels for each sample are learned in a way similar to the mixture of experts. In the spatial domain, to handle object deformations, Deformable Kernels~\cite{gao2019deformable} directly resamples the original kernel space to adapt the effective receptive field (ERF) while leaving the receptive field untouched. Local Relation Networks~\cite{hu2019local} adaptively determine aggregation weights for spatial dimension based on the compositional relationship of local pixel pairs. Non-local~\cite{wang2018non} operation computes the response at each position as the weighted sum of the features at all positions, which can make it capture long-range dependencies. Different from above dynamic methods, apart from dynamic filters, DRConv achieves a dynamic guided mask to automatically determine the distribution of multiple filters so that it can process variable distribution of spatial semantics. However, these methods have high computational and memory complexity which significantly limits the efficiency of the model.

%


\subsection{CNN VS Transformers}
Apart from designing new Vision Transformers, some works focus on exploring the relationship between Transformer and CNN. \cite{zhao2021battle} performs empirical research on various DNN frameworks (e.g., CNN, Transformer, and MLP) in order to grasp their benefits and drawbacks better. \cite{cordonnier2019relationship} shows that self-attention layers can learn to behave similar to convolutional layers. \cite{tay2021pre} discovers that CNN-based pre-trained models are competitive and outperform their Transformer counterparts in some NLP settings. \cite{han2021connection} recasts local attention as a channel-wise locally-connected layer and empirically find that the models based on depth-wise convolution with lower processing complexity perform on par with or somewhat better than Swin Transformer. In this paper, we propose an effective general convolution module, which narrows the gap between CNNs and Transformers.

\section{Method}
\label{sec:Method}

\begin{figure}[h]
\begin{center}
\includegraphics[width=1.0\linewidth]{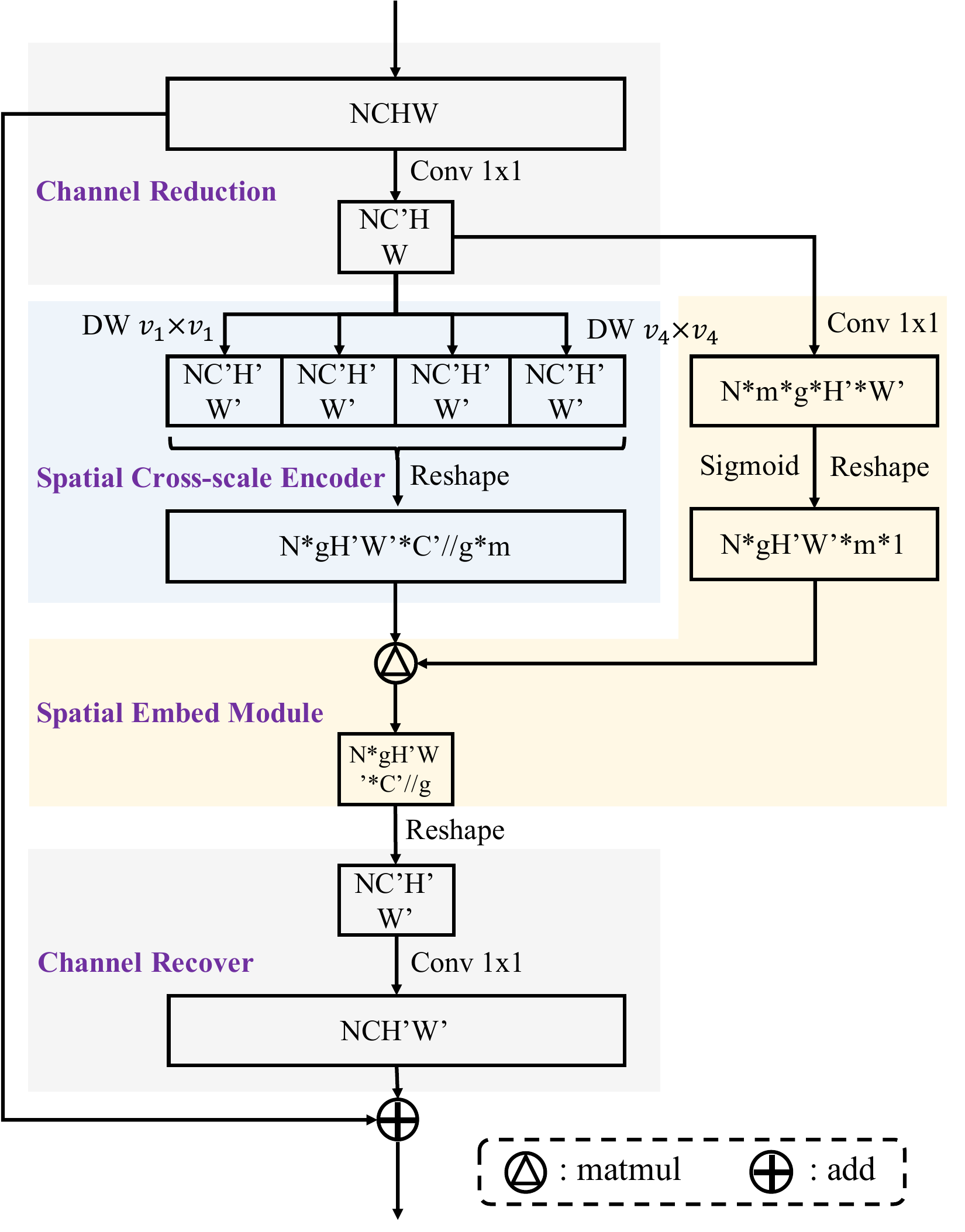}
\vspace*{-8mm}
\end{center}
\caption{Design details of Spatial Cross-scale Convolution Module (SCSC).}
\label{figure:design}
\vspace*{-5mm}
\end{figure}
For CNNs and Transformers, CNNs can exploit large kernels to obtain large receptive fields, and Transformers can get global receptive fields through self-attention. Nevertheless, both large dense kernels and self-attention mechanisms cannot avoid the high computational cost. Moreover, large kernels may hinder the model to capture local features. Therefore, find an efficient way to obtain large effective receptive fields and keep the feature diversity meanwhile attracts increasing attention. 

Since depth-wise convolution is widely used in modern CNNs for its efficiency and effectiveness, it becomes feasible solution to introduce large receptive fields without intensive computational cost. As analyzed in \cite{han2021connection}, depth-wise convolution can resemble local attention (Local Vision Transformer, e.g., Swin-Transformer~\cite{liu2021swin}) in sparse connectivity. But the fixed large windows or kernels may not be a good manner to capture local features and still have space to be improved.

Therefore, based on depth-wise convolution, we applied a wide range of kernel size from $3\times3$ to $13\times13$ to obtain effective receptive fields, so as to capture microscopic and macroscopic feature representation. This way can capture different receptive fields in one layer and strengthen the spatial modeling ability. Furthermore, we put forward an efficient spatial embed module to combine different level spatial features, which can further enhance the overall presentation ability. 

\begin{table*}[t]
\centering
\begin{tabular}{c|c|c|c||c|c}
\hline 
{}   & \tabincell{c}{downsp. rate \\(output size)} & Swin & Swin-SCSC  & ResNet  & ResNet-SCSC     \\
\hline
stage1 
& \tabincell{c}{4× \\(56×56)} 
& \tabincell{l}{concat 4×4, 96-d, LN\\ $\left [\begin{array}{c} \text{win. sz. 7×7} \\ \text{dim 96, head 3} \end{array}\right] \times 2$} 
& \tabincell{c}{concat 4×4, 96-d, LN\\ SCSC block\\ $\left [\begin{array}{c} \text{kernel sz.} \\ \text{[3,11]} \end{array}\right] \times 2$} 
& \tabincell{c}{bottleneck block\\ $\left [\begin{array}{c} \text{kernel sz.} \\ \text{[1,3]} \end{array}\right] \times 3$}
& \tabincell{c}{SCSC block\\ $\left [\begin{array}{c} \text{kernel sz.} \\ \text{[3,9,13]} \end{array}\right] \times 3$} \\
\hline
stage2 
& \tabincell{c}{8× \\(28×28)} 
& \tabincell{l}{concat 2×2, 192-d, LN\\ $\left [\begin{array}{c} \text{win. sz. 7×7} \\ \text{dim 96, head 3} \end{array}\right] \times 2$} 
& \tabincell{c}{concat 2×2, 192-d, LN\\ SCSC block\\ $\left [\begin{array}{c} \text{kernel sz.} \\ \text{[3,9]} \end{array}\right] \times 2$} 
& \tabincell{c}{bottleneck block\\ $\left [\begin{array}{c} \text{kernel sz.} \\ \text{[1,3]} \end{array}\right] \times 4$}
& \tabincell{c}{SCSC block\\ $\left [\begin{array}{c} \text{kernel sz.} \\ \text{[3,7,11]} \end{array}\right] \times 5$} \\
\hline
stage3 
& \tabincell{c}{16× \\(14×14)} 
& \tabincell{l}{concat 2×2, 384-d, LN\\ $\left [\begin{array}{c} \text{win. sz. 7×7} \\ \text{dim 96, head 3} \end{array}\right] \times 6$} 
& \tabincell{c}{concat 2×2, 384-d, LN\\ SCSC block\\ $\left [\begin{array}{c} \text{kernel sz.} \\ \text{[3,7]} \end{array}\right] \times 6$} 
& \tabincell{c}{bottleneck block\\ $\left [\begin{array}{c} \text{kernel sz.} \\ \text{[1,3]} \end{array}\right] \times 6$}
& \tabincell{c}{SCSC block\\ $\left [\begin{array}{c} \text{kernel sz.} \\ \text{[3,5,7]} \end{array}\right] \times 12$} \\
\hline
stage4
& \tabincell{c}{32× \\(7×7)} 
& \tabincell{l}{concat 2×2, 768-d, LN\\ $\left [\begin{array}{c} \text{win. sz. 7×7} \\ \text{dim 96, head 3} \end{array}\right] \times 2$} 
& \tabincell{c}{concat 2×2, 768-d, LN\\ SCSC block\\ $\left [\begin{array}{c} \text{kernel sz.} \\ \text{[3,5]} \end{array}\right] \times 2$} 
& \tabincell{c}{bottleneck block\\ $\left [\begin{array}{c} \text{kernel sz.} \\ \text{[1,3]} \end{array}\right] \times 3$}
& \tabincell{c}{SCSC block\\ $\left [\begin{array}{c} \text{kernel sz.} \\ \text{[3,5]} \end{array}\right] \times 3$} \\
\hline
\end{tabular}
\caption{ Detailed architecture specifications. The kernel size (sz.) of our SCSC can be any combination for the specific tasks, here we just give some examples.}
\label{table:Architecture}
\vspace{-1.2em} 
\end{table*}

\subsection{SCSC: Spatial Cross-scale Convolution Module}
For clear illustration, we split SCSC into four steps: Channel Reduction, Spatial Cross-scale Encoder, Spatial Combination, and Channel Recover. Taking a $K$ layers CNN for example, the input of the $k_{th}$ layer can be denoted as $X^k \in \mathbb{R}^{N \times C \times H \times W}$, where N, C, H, W are batchsize, channel, height, and width respectively. As shown in Figure~\ref{figure:design}, for Channel Reduction, we use conv $1\times1$ as the mapping function,
\begin{eqnarray}
	X^k_d = W^k_d \otimes X^k , 
\label{Equation:Eq1}
\end{eqnarray}
where $\otimes$ denotes convolutional operation, $X^k_d \in \mathbb{R}^{N \times C//m \times H \times W}$, m is a constant and means how many different kernels in Spatial Cross-scale Encoder. Then we calculate the Spatial Cross-scale Encoder by
\begin{eqnarray}
X^k_s =\left\{X^k_{s_1}, ..., X^k_{s_i}, ..., X^k_{s_m} \right\},
\label{Equation:Eq2}
\end{eqnarray}
\begin{eqnarray}
X^k_{s_i} = W^k_{s_i} \otimes X_d^k \ ,\  W^k_{s_i} \in \mathbb{R}^{ C//m \times C//m \times v \times v}  ,
\label{Equation:Eq3}
\end{eqnarray}
in which $i\in [1,m], v\in [3,13], X^k_{s_i} \in \mathbb{R}^{N \times C//m \times H' \times W'}$. After getting different level spatial features, we design a Spatial Embed Module to dynamically fuse multi-scale features at each spatial position (i.e., pixel level).
\begin{eqnarray}
	\mathcal{M}^k = SEM(X^k_d),
\label{Equation:Eq4}
\end{eqnarray}
in $SEM( )$, we use $1 \times 1$ conv to reduce the channel number to $m*g$, then apply sigmoid function, the result noted as $\mathcal{M}^k \in \mathbb{R}^{ N \times m*g \times H' \times W'}$, $g$ is a constant and means the combination times, which can bring more dynamic combinations. We reshape the $X^k_s$ to $N \times g*H'*W' \times C//m//g \times m$, and $\mathcal{M}^k$ to $ N \times g*H'*W' \times m \times 1$. and then we can get the embedded feature,
\begin{eqnarray}
	X^k_e = Matmul(X^k_s, \mathcal{M}^k), 
\label{Equation:Eq5}
\end{eqnarray}
where $X^k_e \in \mathbb{R}^{N \times C//m \times H' \times W'}$. 
\cxj{}

Finally, we product Channel Recover operation by
\begin{eqnarray}
	X^k_o = W^k_r \otimes X^k_e ,
\label{Equation:Eq6}
\end{eqnarray}
and $X^k_o \in \mathbb{R}^{N \times C \times H' \times W'}$. All the operations are differentiable, so our objective function is 
\begin{eqnarray}
	\min\limits_{W^k_d,W^k_s, W_{\mathcal{M}^k}, W^k_r|_{k=1}^K}\frac{1}{n}\sum_{i=1}^{n}\mathcal{L}(Y_i;\hat{Y}|X_i),
	\label{Equation:Eq9}
\end{eqnarray}
in which $W$ is the learnable weights, $(X_i, Y_i), i\in[1,n]$ is the input image and label, $\hat{Y}$ is the predicted label. $\mathcal{L( )}$ is the loss function.

\subsection{Architecture}
The proposed SCSC module can be easily embedded in any existing architecture. As shown in Table~\ref{table:Architecture}, taking the most used CNN ResNet for example, we directly replace the original bottleneck block of ResNet-50 with our SCSC block. For Transformer, we replace the self-attention with the proposed SCSC module, in which the pre-linear-projections and post-linear-projections over the values can be regarded as the 1×1 convolutions in our design. 

\begin{table*}[ht]
\centering
\begin{tabular}{c|ccccc}
\hline 
Model   & Image Resolution & Kernel Size Range& Params  & FLOPs  & Top-1 Acc. (\%)     \\
\hline
ResNet-50 \cite{he2016deep}             & $224\times224$ & $1\times1$ to $3\times3$ & 26M & 4.1G  & 76.2 \\
ResNet-101 \cite{he2016deep}            & $224\times224$ & $1\times1$ to $3\times3$ & 45M & 7.9G  & 77.4 \\
ResNet-152 \cite{he2016deep}            & $224\times224$ & $1\times1$ to $3\times3$ & 60M & 11.6G & 78.3 \\
SKNet-50 \cite{li2019selective}         & $224\times224$ & $1\times1$ to $5\times5$ & 27M & 4.5G  & 79.3 \\
SKNet-100 \cite{li2019selective}        & $224\times224$ & $1\times1$ to $5\times5$ & 49M & 8.5G  & 79.8 \\
SENet  \cite{hu2018squeeze}             & $224\times224$ & $1\times1$ to $3\times3$ & 22M & 3.9G  & 79.9 \\
ResNeXt \cite{xie2017aggregated}        & $224\times224$ & $1\times1$ to $3\times3$ & 25M & 4.2G  & 80.1 \\
\hline                             
ResNet-SCSC-V1 (Ours)                      & $224\times224$ & $1\times1$ to $13\times13$ & \textbf{10M} & \textbf{1.7G}  & \textbf{79.4} \\
ResNet-SCSC-V2 (Ours)                      & $224\times224$ & $1\times1$ to $13\times13$ & \textbf{12M} & \textbf{2.2G}  & \textbf{80.3} \\
ResNet-SCSC-V3 (Ours)                      & $224\times224$ & $1\times1$ to $13\times13$ & \textbf{25M}  & \textbf{4.5G}   & \textbf{81.5} \\
\hline 
ViT-B/16 \cite{dosovitskiy2020image}    & $384\times384$ & Global Receptive Field     & 86M  & 55.4G  & 77.9 \\
ViT-L/16 \cite{dosovitskiy2020image}    & $384\times384$ & Global Receptive Field     & 307M & 190.7G & 76.5 \\
DeiT-S \cite{touvron2021training}       & $224\times224$ & Global Receptive Field     & 22M  & 4.6G   & 79.8 \\
DW-Conv.-T \cite{han2021connection}   & $224\times224$ & $7\times7$                 & 24M  & 3.8G   & 81.3 \\
Swin-T \cite{liu2021swin}             & $224\times224$ & Local Window $7\times7$    & 28M  & 4.5G   & 81.3 \\
ConvNet-T \cite{liu2022convnet}        & $224\times224$ &  $7\times7$    & 28M  & 4.5G   & 82.1 \\
\hline
Swin-T-SCSC (Ours)                      & $224\times224$ & $1\times1$ to $11\times11$ & \textbf{22M}  & \textbf{3.5G}  & \textbf{81.6} \\
ConvNet-T-SCSC (Ours)       & $224\times224$ & $1\times1$ to $11\times11$    & 28M  & 4.5G   & \textbf{82.2} \\
\hline
\end{tabular}
\caption{ Comparison of Top-1 classification accuracy with different architectures (CNNs and Transformers) and some state-of-the-art backbones on ImageNet. }
\vspace*{-5mm}
\label{table:ImageNet}
\end{table*}

\subsection{Discussion and Limitation}
Some existing methods have contributed to exploring the effectiveness of multi-scale features, but they are suboptimal in terms of efficiency. For example, PSP~\cite{he2015spatial} and ASPP~\cite{chen2017deeplab} use sampling operators (e.g., pooling or atrous convolution) to efficiently extract multi-scale features, but this comes with the expense of losing spatial information. CrossFormer and GoogleNet use resource-intensive operations, such as regular convolution with large kernels, to extract multi-scale features, and they fuse them through static concatenation. However, this approach significantly increases model parameters.

In contrast, our SCSC approach not only introduces a wide range of receptive fields for effective representation learning but also incorporates weight sharing and dynamic mechanisms for greater efficiency. This allows our model to achieve high performance while keeping the model size and computation cost low. 

In details, first, our SCSC is compact and effective,  the Channel Reduction step decreases the size of the input feature maps, and then inputs feature maps into the Spatial Cross-scale Encoder, which consists of depth-wise convolution with a wide range of kernel size to model the spatial information. This manner can effectively avoid the intensive computational cost. In the meanwhile, the wide range of kernel size provides different receptive fields in one layer, small kernels for the detailed local information, large kernels for the semantic dependence. Furthermore, we design a dynamic Spatial Embed Module to merge the different spatial information. The proposed SCSC naturally aggregates the advantage of CNN and Transformer, small kernels and weight sharing from the CNN, large receptive fields and dynamic from the Transformer. However, it still has its limitation: depth-wise convolution may need exceptional acceleration in practice. 


\section{Experiments}
In this section, we evaluate the effectiveness of our proposed SCSC module by embedding it into the classical CNN backbones (ResNet-50~\cite{he2016deep}, MobileFaceNet~\cite{chen2018mobilefacenets} and FaceResNet~\cite{ding2021repmlp}),  and state-of-the-art Swin-Transformer~\cite{liu2021swin} respectively. We conduct experiments for SCSC-based architectures on ImageNet~\cite{russakovsky2015imagenet}, MS1M-V2~\cite{guo2016ms}, COCO 2017~\cite{lin2014microsoft} and ADE20K\cite{zhou2019semantic} in terms of image classification, face recognition, object detection and segmentation.

\subsection{Classification}
\label{Sec:Classification}

We take classical ResNet-50~\cite{he2016deep} and state-of-the-art Swin-Transformer~\cite{liu2021swin} as the backbone to evaluate SCSC by replacing their original components respectively. 

\textbf{Settings:} The ImageNet 2012 dataset~\cite{russakovsky2015imagenet} is a well-known image classification dataset includes $1.28$ million training images and 50k validation images from 1000 classes. All our models are trained on the whole training dataset and validated using the single-crop top-1 validation accuracy. The training settings follow \cite{liu2021swin}, all models in our experiments are trained for 300 epochs with AdamW~\cite{kingma2014adam} optimizer and a cosine decay learning rate scheduler starts from 0.001/0.0005. We set batch size as 1024 and weight decay as 0.05.  

\textbf{Resnet50 with SCSC.} We directly replace the original bottleneck block with our SCSC block. Following the mainstream works, we refer to the four residual stages of ResNet-50 as c2, c3, c4, c5, respectively. And the block number for the four stages are 3, 4, 6, 3, respectively. For ResNet-SCSC-V1, the four stages' input/output channel number are 96, 192, 384, 512, expansion is set as 2. The kernel sets for the four stages are [3,9,13], [3,7,11], [3,5,7] and [3,5] respectively. We increase the numbers of blocks in c2,c3,c4,c5 from 3,4,6,3 to 3,4,8,3 so that the FLOPs can match the original network. For ResNet-SCSC-V2, most setting keep the same as ResNet-SCSC-V1, except increasing the numbers of blocks in c2,c3,c4,c5 from 3,4,8,3 to 3,5,12,3. For ResNet-SCSC-V3, most settings keep the same as ResNet-SCSC-V2, except the expansion is set as 3. We list the architecture of ResNet-SCSC-V3 in Table~\ref{table:Architecture}.

\textbf{Swin-Transformer with SCSC.} We replace local self-attention with our SCSC module.  The pre-linear-projections and post-linear-projections over the values can be regarded as the 1×1 convolutions in our design. 

The results are shown in Table~\ref{table:ImageNet}. As can be seen, for ResNet-SCSC-V1 and ResNet-SCSC-V2, we use much less computational cost and parameters to obtain competitive results. For ResNet-SCSC-V3, with comparable model complexity, we successfully make the performance of a classical CNN architecture (ResNet) match the Transformers' performance.  For Swin-Transformer, with our SCSC module, it achieves a higher accuracy with 1G (22\%) FLOPs fewer computational cost. These results show that SCSC-based architectures not only have a considerable improvement, but also saving computational cost and parameters, demonstrating the effectiveness of our method.

\subsection{Face Recognition}
We further evaluate the effectiveness of our SCSC on Face Recognition task. MobileFaceNet~\cite{chen2018mobilefacenets}  and FaceResNet~\cite{ding2021repmlp}  are applied as our backbone with input size $96 \times 96$. 

\noindent\textbf{Settings:} MS1M-V2 dataset is a large-scale face dataset with 5.8M images from 85k celebrities. We use a refined semi-automatic version of the MS-Celeb-1M dataset~\cite{guo2016ms} which consists of 1M photos from 100k identities for training. The dataset we use for validation is MegaFace~\cite{kemelmacher2016megaface}, which includes 1M images of 60k identities as the gallery set and 100k images of 530 different individuals. We use SGD with a momentum of 0.9 to optimize the model, and the batch size is 512. We train all the models for 420k iterations. The learning rate begins with 0.1 and is divided by 10 at 252k, 364k, and 406k iterations. For evaluation, we use face identification metric which refers to the rank-1 accuracy on MegaFace as the evaluation indicator.

\noindent\textbf{MobileFaceNet with SCSC.} We directly replace the original bottleneck block with our SCSC block. In \cite{chen2018mobilefacenets}, there are 5 stages, we denote them as s1, s2, s3, s4, s5, respectively. And the block number for the five stages are 5, 1, 6, 1,2, respectively. For MobileFaceNet-SCSC, the kernel sets for the five stages are [3,9], [3,7], [3,7], [3,5] and [3,5] respectively. The kernel size is relatively small because of the input image size is small.  We set expansion as 3 so that not reducing too many FLOPs since MobileFaceNet is already an efficient network.

\noindent\textbf{FaceResNet with SCSC.} As in FaceResNet~\cite{ding2021repmlp}, we denote the four residual stages of FaceResNet as c2, c3, c4, c5, respectively. And the block number for the four stages are 3, 2, 2, 2. For FaceResNet-SCSC, We set the numbers of blocks in c2,c3,c4,c5 as 6,6,6,4. The kernel sets for the four stages are [5,11], [3,9], [3,5] and [3,3] respectively. 

\begin{table}[t]
\centering
\begin{tabular}{cccc}
\hline 
Model   & Params & FLOPs & Acc.(\%)    \\
\hline
MobileFaceNet~\cite{chen2018mobilefacenets}         & 0.98M & 162M  & 90.9 \\
MobileFaceNet-SCSC                                  & \textbf{0.89M} & \textbf{146M}  & \textbf{92.0} \\
FaceResNet~\cite{ding2021repmlp}                    & 40.3M & 1050M & 92.9 \\
FaceResNet-SCSC                                     & \textbf{8.3M}  & \textbf{330M}  & \textbf{95.6} \\
\hline
\end{tabular}
\caption{Results of SCSC on Megaface. ``Acc.'' refers to the rank-1 face identification accuracy with 1M distractors.}
\vspace{-1.2em} 
\label{table:Face_Recongnition}
\end{table}

As Table~\ref{table:Face_Recongnition} shows, given that MobileFaceNet is already an efficient network, MobileFaceNet-SCSC still outperforms the baseline by 1.1\% gain even with fewer FLOPs and parameters. Moreover, FaceResNet-SCSC surpasses FaceResNet by a large margin of 2.7\% accuracy with 68\% fewer computational cost and 79\%  fewer parameters, further indicating the superiority of our proposed SCSC.

\begin{table}[t]
\addtolength{\tabcolsep}{-3pt}
\centering
\begin{tabular}{c|cc|cc}
\hline
Backbone        & box AP & mask AP & FLOPs & Params \\ \hline
R50~\cite{he2016deep}        & 38.2   & 34.7    & 260.1G    & 44.2M       \\
R101~\cite{he2016deep}       & 40.0   & 36.1    & 336.2G     & 63.2M       \\
X101-32x4d~\cite{xie2017aggregated} & 41.9   & 37.5    & 340.0G     & 62.8M       \\
X101-64x4d~\cite{xie2017aggregated} & 42.8   & 38.4    & 493.4G     & 101.9M      \\
Swin-T~\cite{liu2021swin}          & 42.7   & 39.3    & 263.8G     & 47.8M       \\ \hline
ResNet-SCSC          & \textbf{44.0}   & \textbf{40.4}    & 270.8G    & 44.9M       \\
Swin-SCSC       & \textbf{43.2 }  & \textbf{39.6}    & \textbf{240.5G}     & \textbf{41.8M}      \\ \hline
\end{tabular}
\caption{Results of object detection and instance segmentation on the COCO \textit{mini-val} with Mask R-CNN (1x schedule). FLOPs are measured on an 800 $\times$ 1280 image.}
\vspace{-.3em} 
\label{table:det1x}
\end{table}

\begin{table}[t]
\addtolength{\tabcolsep}{-3pt}
\centering
\begin{tabular}{c|cc|cc}
\hline
Backbone     & box AP & mask AP & FLOPs & Params \\ \hline
R50~\cite{he2016deep}          & 46.3   & 43.4    & 739G  & 82M      \\
DeiT-S~\cite{touvron2021training}       & 48.0   & 41.4    & 889G  & 80M      \\
Swin-T~\cite{liu2021swin}       & 50.4   & 43.7    & 742G  & 86M      \\
DW Conv.-T~\cite{han2021connection}   & 49.9   & 43.4    & 730G  & 82M      \\
 \hline
ResNet-SCSC       & 50.3   & \textbf{43.7}   & 749G  & 83M      \\
Swin-SCSC    & 49.9   & 43.2    & \textbf{719G}  & \textbf{80M}     \\ \hline
\end{tabular}
\caption{Results of object detection and instance segmentation performance on the COCO \textit{mini-val} with Cascade Mask R-CNN (3x schedule). FLOPs are measured on an 800 $\times$ 1280 image.}
\vspace{-.3em} 
\label{table:det3x}
\vspace*{-2mm}
\end{table}

\subsection{Detection}

To evaluate the effectiveness of our SCSC on object detection, we utilize the COCO 2017 dataset~\cite{lin2014microsoft} which consists of 80k train images and 40k val images. 

\noindent\textbf{Settings:} we exploit Mask R-CNN~\cite{he2017mask} and Cascade Mask R-CNN~\cite{cai2018cascade} framework with FPN~\cite{lin2017feature}.  
For Mask-RCNN, the implementation is based on MMDetection~\cite{chen2019mmdetection}. We train the detection networks on 8 GPU with mini-batch 2 per GPU for 1x schedule (12 epochs). We use AdamW optimizer, and the initial learning rate is 0.0001, started after 500 iteration warmup and decayed by 0.1 times at the 8th and 11th epoch. We use stochastic drop path regularization of 0.2 and weight decay of 0.05. All other hyper-parameters follow the default settings in MMDetection~\cite{chen2019mmdetection}. For Cascade Mask-RCNN, we follow the implementation, training, and test settings from Swin Transformer~\cite{liu2021swin}. Backbone weights are initialized by the parameters of Swin-SCSC and ResNet-SCSC respectively. 

 As shown in Table~\ref{table:det1x}, we compare Swin-SCSC and ResNet-SCSC with standard ConvNets, i.e., ResNe(X)t, and previous Transformer networks, e.g., Swin-T with Mask R-CNN. The comparisons are conducted by changing only the backbones with other settings unchanged. Our ResNet-SCSC architecture achieves 5.0\% improvement over baseline ResNet-50. Compared with strong baseline Swin-T, ResNet-SCSC improves 1.3\%  box AP and 1.1\% mask AP over Swin-T. Swin-SCSC achieves better performance even with $23G$ FLOPs fewer computational cost and $6M$ fewer parameters, illustrating that SCSC also works on downstream tasks.


\subsection{Semantic Segmentation}


\begin{table}[]
\centering
\begin{tabular}{c|c|cc}
\hline
Backbone     & mIoU  & FLOPs & Params \\ \hline
R50~\cite{he2016deep}          & 42.1  & 952G  & 67M      \\
R101~\cite{he2016deep}         & 43.8  & 1029G & 86M      \\
DeiT-S~\cite{touvron2021training}       & 42.9  & 1099G & 52M      \\
DW Conv.-T~\cite{han2021connection}   & 45.5  & 928G  & 56M      \\
Swin-T~\cite{liu2021swin}       & 44.4 & 941G  & 60M      \\ \hline
ResNet-SCSC       & \textbf{45.7} & 956G  & 64M      \\
Swin-SCSC    & 44.0 & \textbf{916G}  & \textbf{54M}      \\\hline
\end{tabular}
\caption{Results of semantic segmentation on the ADE20K \textit{val} set with UperNet. FLOPs are measured on an 512 $\times$ 2048 image.}
\vspace{-.2em} 
\label{table:seg}
\vspace{-3mm}
\end{table}

{\bf Settings:} ADE20K\cite{zhou2019semantic} is a widely used semantic segmentation dataset, covering a broad range of 150 semantic categories. It has 25K images, with 20K for training, 2K for validation, and another 3K for testing. We utilize the implementation of UperNet \cite{xiao2018unified} in MMSegmentation \cite{mmseg2020} as our base framework for its high efficiency. We use the same setting as the Swin Transformer~\cite{liu2021swin}.
Models are trained on 8 GPUs with mini-batch 2 per GPU for 160k iterations. We employ the AdamW optimizer with an initial learning rate of 0.00006, a weight decay of 0.01, a scheduler that uses linear learning rate decay, and a linear warmup of 1,500 iterations.  SyncBN and stochastic depth with the ratio of 0.3 is applied for Swin-SCSC.  The experimental results are reported as single-scale testing.  All other hyper-parameters follow the default settings in MMSegmentation~\cite{mmseg2020}.

Table~\ref{table:seg} lists the mIoU, FLOPs and model size (number of parameters) for different backbones. Compared with Swin-T, our ResNet-SCSC achieves +1.3 mIoU higher (45.7 \textit{vs.} 44.4) than Swin-T with slightly larger model size and computation cost. Our Swin-SCSC achieves comparable performance with 25G less computation cost and 6M fewer parameters.

\subsection{Ablation Study}


We conduct a series of experiments to evaluate the impact of specific design choices in our spatial Cross-scale Convolution Module. We use the ImageNet classification dataset~\cite{russakovsky2015imagenet} for our experiments, and follow the same experimental settings as described in section~\ref{Sec:Classification} by default. 
%

\noindent\textbf{The effectiveness of combination number $g$ in Spatial Embed Module:} We use ResNet-SCSC to explore which combination number $g$ in Spatial Embed Module is better. We set $g$ as 2, 4, 8, respectively, and the Top-1 accuracy is 81.1\%,  81.5\%, and 81.2\% respectively. Therefore, we set $g=4$ by default. 


\noindent\textbf{The Effect of Spatial Embed Module:} We design the Spatial Embed Module (SEM) for fusing multi-scale features dynamically. For ResNet-SCSC, if we only use 7x7 kernel size, we will get 80.6\% top-1 accuracy. Adding Spatial Cross-scale Encoder, the accuracy can reach 81.0\%. Further adding Spatial Embed Module, we can obtain 81.6\% accuracy. This shows the effectiveness of dynamic fusing by using our SEM.

\begin{figure}[h]
\begin{center}
\vspace*{-5mm}
\includegraphics[width=1.0\linewidth]{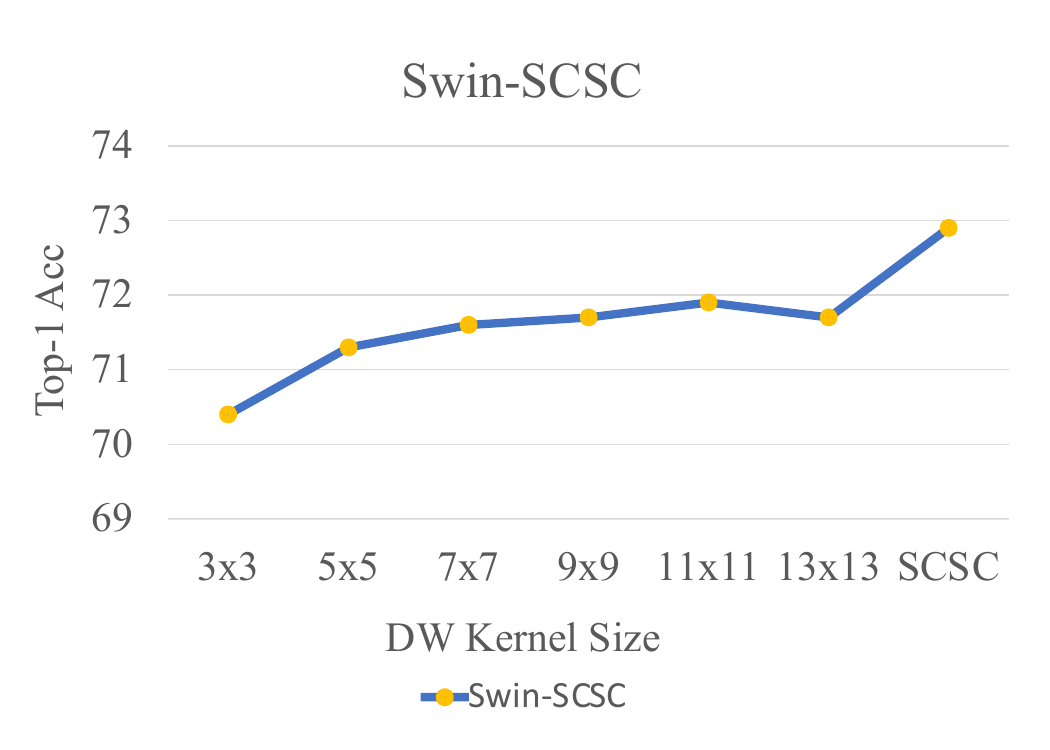}
\vspace*{-10mm}
\end{center}
\caption{Influence of DW kernel size in Spatial Embed Module. We set DW kernel size as $3\times3$, $5\times5$, $7\times7$, $9\times9$, $11\times11$, $13\times13$, SCSC setting,  respectively. Larger kernel will result better performance, but when the kernel size increase too large, the accuracy begin to decline. Our SCSC achieves a large improvement by using different receptive fields in one layer}
\label{Fig:Ablation}
\end{figure}


\noindent\textbf{Kernel size:} We train Swin-SCSC for 100 epochs to explore the choices of kernel sizes. We set the size of depth-wise (DW) kernels with $3\times3$, $5\times5$, $7\times7$, $9\times9$, $11\times11$, $13\times13$, as well as our SCSC setting. As shown in Figure~\ref{Fig:Ablation}, larger kernels results in better accuracy but when the kernel size exceeds $11\times11$, accuracy begins to decline. Our results demonstrate that both excessively small and excessively large kernel sizes are suboptimal. Furthermore, using different receptive fields in one layer achieves the highest accuracy and significantly outperforms other methods. This suggests that utilizing multiple receptive fields can effectively enhance the performance of model.

\section{Conclusion and Future Work}

Different from the mainstream CNN's small kernel size and the Transformer's global receptive field, we increase the spatial modeling ability by using a wide range of kernel sizes. Furthermore, we design an efficient spatial embedding module to merge the different spatial representation features. As a result, the proposed SCSC naturally combines the advantage of CNNs and Transformers, small kernel and weight sharing from the CNN, large receptive field and dynamic from the Transformer. Intensive experiments illustrate the effectiveness and generalization of our SCSC. In the future, we will explore more about the relationship between CNNs and Transformers.


{\small
\bibliographystyle{ieee_fullname}
\bibliography{egbib}
}

\end{document}